\documentclass[conference]{IEEEtran}
\IEEEoverridecommandlockouts
\usepackage{cite}
\usepackage{amsmath,amssymb,amsfonts}
\usepackage{algorithmic}
\usepackage{graphicx}
\usepackage{textcomp}
\usepackage{xcolor}
\usepackage{colortbl}
\usepackage{placeins}
\usepackage{booktabs}
\usepackage{stfloats}
\usepackage{url} 
\definecolor{best}{RGB}{255,200,200}
\definecolor{second}{RGB}{200,255,208}

\def\BibTeX{{\rm B\kern-.05em{\sc i\kern-.025em b}\kern-.08em
    T\kern-.1667em\lower.7ex\hbox{E}\kern-.125emX}}
\begin{document}

\title{HiCrew: Hierarchical Reasoning for Long-Form Video Understanding via Question-Aware Multi-Agent Collaboration}


\author{
\IEEEauthorblockN{Yuehan Zhu\textsuperscript{*}, Jingqi Zhao\textsuperscript{*}, Jiawen Zhao\textsuperscript{*}, Xudong Mao, Baoquan Zhao\textsuperscript{\dag}}
\\
\IEEEauthorblockA{\textit{School of Artificial Intelligence, Sun Yat-sen University, China}
\thanks{$^*$ Equal contribution.}
\thanks{$^\dag$ Corresponding author. zhaobaoquan@mail.sysu.edu.cn}
}
\thanks{This work was supported by the National Natural Science Foundation of China under Grants 61902087 and 62302535.}
}

\maketitle

\begin{abstract}
Long-form video understanding remains fundamentally challenged by pervasive spatiotemporal redundancy and intricate narrative dependencies that span extended temporal horizons. While recent structured representations compress visual information effectively, they frequently sacrifice temporal coherence, which is critical for causal reasoning. Meanwhile, existing multi-agent frameworks operate through rigid, pre-defined workflows that fail to adapt their reasoning strategies to question-specific demands. In this paper, we introduce HiCrew, a hierarchical multi-agent framework that addresses these limitations through three core contributions. First, we propose a Hybrid Tree structure that leverages shot boundary detection to preserve temporal topology while performing relevance-guided hierarchical clustering within semantically coherent segments. Second, we develop a Question-Aware Captioning mechanism that synthesizes intent-driven visual prompts to generate precision-oriented semantic descriptions. Third, we integrate a Planning Layer that dynamically orchestrates agent collaboration by adaptively selecting roles and execution paths based on question complexity. Extensive experiments on EgoSchema and NExT-QA validate the effectiveness of our approach, demonstrating strong performance across diverse question types with particularly pronounced gains in temporal and causal reasoning tasks that benefit from our hierarchical structure-preserving design. Our codebase is available at \url{https://github.com/SYSUzzz/HiCrew}.
\end{abstract}


\section{Introduction}
\label{sec:intro}

The proliferation of long-form video content has positioned video understanding as a pivotal challenge in multimedia. Unlike short clips, long-form videos exhibit substantial spatiotemporal redundancy and complex causal dependencies that demand sophisticated reasoning. Recent advances explore extended context windows of large language models (LLMs) \cite{b1, b2, b3}, structured representations \cite{b4, b5}, and multi-agent frameworks \cite{b6}, yet each faces substantial limitations.

\begin{figure}[!t]
\centerline{\includegraphics[width=\columnwidth]{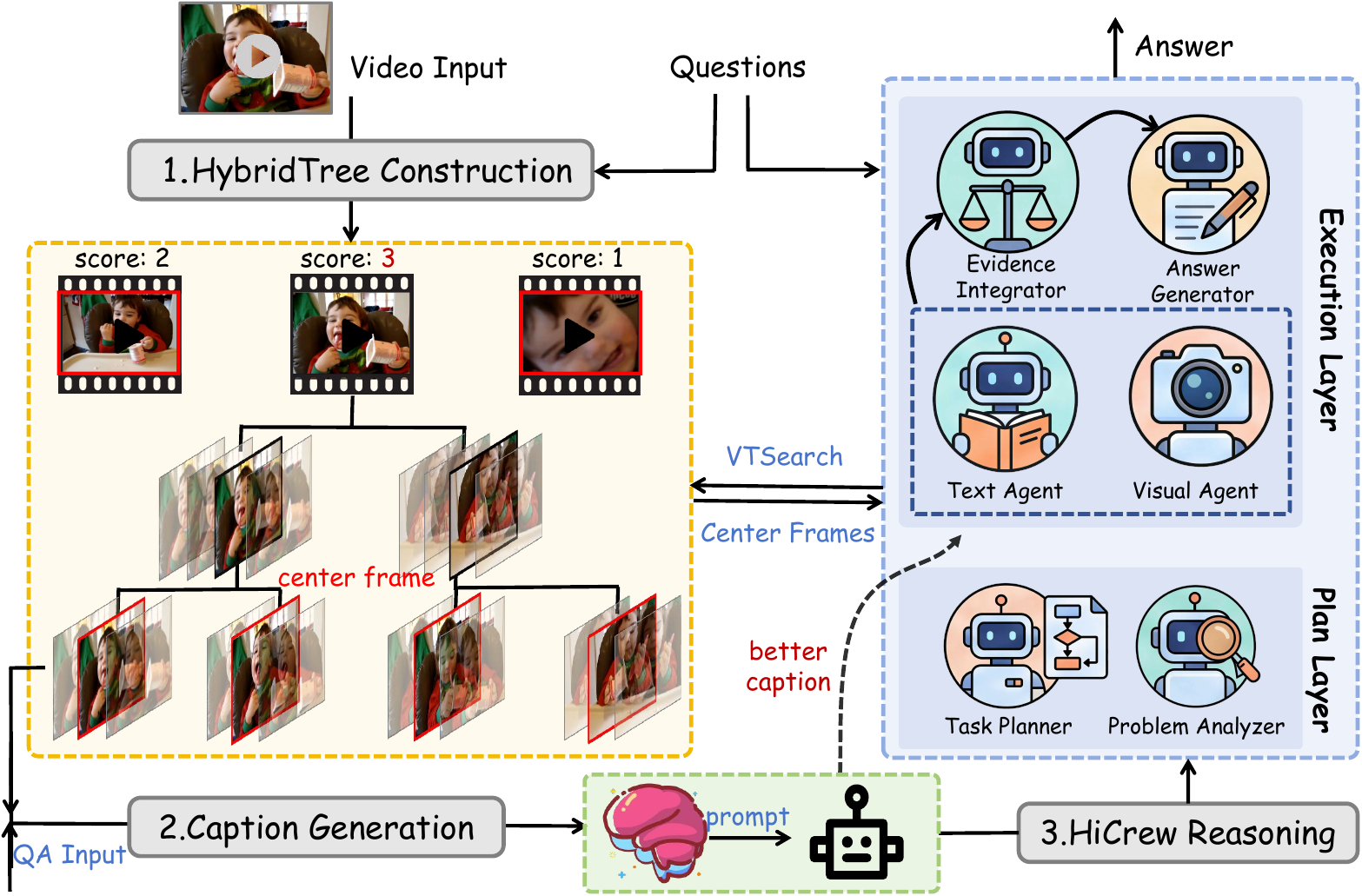}}
\caption{Overview of the proposed HiCrew framework.}
\label{fig1}
\end{figure}

Processing long-form videos within limited context windows has motivated extensive research into efficient representation. Early approaches like LongVA \cite{b8} extended visual encoder sequence lengths but suffered from poor scalability, prompting a shift toward structured representation. Hierarchical structures have emerged as powerful tools for organizing video content at multiple granularities. HierVL \cite{b20} pioneered bottom-up hierarchical video-language embeddings, while VideoTree \cite{b4} introduced hierarchical clustering \cite{b21} achieving competitive performance without training. However, global clustering groups temporally distant frames, obscuring causal and temporal relationships essential for reasoning.

Complementary compression methods distill videos into compact representations while preserving semantics. BREASE \cite{b9} draws inspiration from human episodic memory, VideoMamba \cite{b10} leverages State Space Models for temporal modeling, and BIMBA \cite{b5} extends this through bidirectional selective scanning. However, aggressive compression may eliminate precise visual details needed for accurate reasoning.

Unlike monolithic models, agents achieve deeper interpretation by orchestrating multimodal tools and external knowledge \cite{b2, b3, b11, b12}. Early single-agent Multimodal LLMs for video question answering lacked iterative refinement \cite{b13}, which motivated Retrieval-Augmented Generation (RAG) and collaborative multi-agent systems \cite{b6}. However, existing frameworks still rely on static collaboration with pre-defined sequences and uniform strategies. Consequently, simple queries incur unnecessary computational costs while complex ones receive insufficient reasoning depth, highlighting the absence of adaptive planning as a critical limitation.

To address these challenges, we propose HiCrew, a hierarchical multi-agent framework that reconciles structural efficiency with temporal integrity through three complementary designs (Fig.~\ref{fig1}). First, our Hybrid Tree structure preserves temporal topology through shot boundary detection initialization, conducting hierarchical expansion exclusively within shots rather than performing global clustering. High-relevance shots undergo deep refinement while low-relevance segments remain coarse-grained. Second, Question-Aware Captioning analyzes question intent to synthesize targeted visual prompts, directing vision-language models to question-relevant elements. Third, our Planning Layer performs question-aware orchestration, dynamically determining minimal agent subsets and constructing question-specific workflows. By analyzing question semantics to determine optimal agent composition, our framework achieves adaptive reasoning tailored to question demands.
The contributions of this work are as follows:
\begin{itemize}
    \item We propose Hybrid Tree construction that preserves temporal structure while achieving effective visual compression through selective hierarchical clustering, addressing the limitations of global clustering methods that disrupt narrative coherence. 
    \item We develop Question-Aware Captioning that actively generates intent-driven visual guidance for precision-oriented semantic description.
    \item We design HiCrew, a hierarchical multi-agent system with dynamic planning capabilities that achieves strong and consistent performance on established benchmarks while demonstrating enhanced adaptability across diverse question types.
\end{itemize}

\section{Proposed Method}

\begin{figure*}[!t]
\centering
\includegraphics[width=\textwidth]{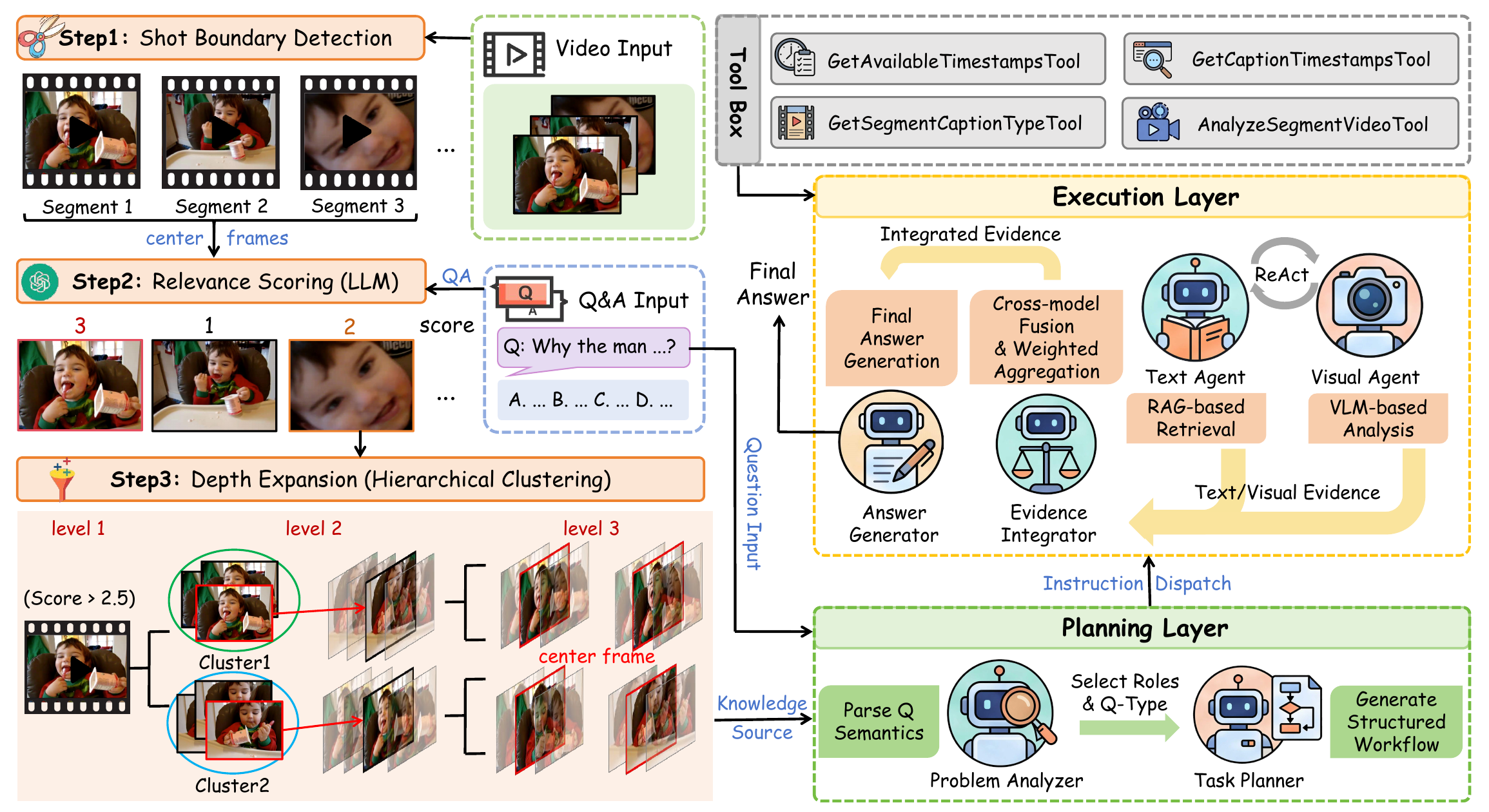}
\caption{Architecture of HiCrew framework showing Hybrid Tree construction (left) and hierarchical multi-agent collaboration (right).}
\label{fig:2}
\end{figure*}

Fig.~\ref{fig:2} illustrates the HiCrew architecture comprising video preprocessing (left panel) and hierarchical multi-agent collaboration (right panel). We detail three key components in the following subsections: Hybrid Tree construction (Section II-A), Question-Aware Captioning (Section II-B), and multi-agent collaboration with adaptive planning (Section II-C).

\subsection{Hybrid Tree Construction}

To reconcile effective visual compression with strict temporal preservation, we introduce the Hybrid Tree structure, $T_{hyb}$, which maintains the video's narrative topology while condensing redundant information.
Unlike prior structured methods that perform global frame clustering, we conceptualize the video $V$ as a composition of temporally contiguous shots. We employ shot boundary detection to segment the video into an ordered sequence $S = \{s_1, s_2, \ldots, s_N\}$, where each shot $s_i$ represents a semantically coherent visual segment. These shots form the first-layer nodes of $T_{hyb}$, establishing an invariant temporal foundation. This initialization strategy ensures that all subsequent compression operations preserve the video's narrative structure.

We introduce a lightweight LLM-based scoring mechanism $f_{score}$ to identify question-relevant content. For each shot $s_i$, we extract a representative frame $f_i$  (selected as the frame nearest to the clip's visual centroid) along with its corresponding textual caption $c_i$. Given the user question $Q$, we compute a relevance score:
\begin{equation}
score_i = f_{score}(f_i, c_i, Q).
\end{equation}
This score quantifies the semantic alignment between the shot content and the question, enabling informed decisions about where to allocate representational capacity.

For shots with high relevance scores ($score_i > \tau$), we hypothesize that they contain critical visual details necessary for accurate question answering. Within these high-relevance shots, we perform K-Means hierarchical clustering using CLIP features to generate finer-grained sub-event nodes, thereby increasing the tree depth and capturing subtle visual distinctions. Conversely, shots with low relevance are retained as leaf nodes without further expansion, avoiding the introduction of irrelevant noise into the representation. This selective expansion strategy achieves an optimal balance: global temporal structure remains intact through shot-level organization, while local detail is enhanced in question-critical regions.

\subsection{Question-Aware Captioning}

Generic video captions, while descriptive, often lack the specificity required to support precise reasoning about particular questions. We address this limitation through an intent-driven caption generation mechanism.
Given a video $V$ and its associated question set $\mathcal{Q}_V$, our system first performs semantic classification, categorizing questions into three fundamental types: Causal ($\mathcal{C}_{cau}$), which focus on action motivations and methods; Temporal ($\mathcal{C}_{tem}$), which concern event sequencing and timing; and Descriptive ($\mathcal{C}_{des}$), which address scene attributes and object properties. 
Subsequently, an LLM functions as a meta-prompt generator, analyzing the characteristic information demands of each question type to synthesize a targeted visual guidance prompt $P_{t}$:
\begin{equation}
P_{t} = \text{LLM}(\mathcal{Q}_V^{t}, \mathcal{T}_{t}), \quad t \in \{\mathcal{C}_{cau}, \mathcal{C}_{tem}, \mathcal{C}_{des}\}
\end{equation}
where $P_{t}$ is the visual prompt, $\text{LLM}$ is the generator, $\mathcal{Q}_V^{t}$ denotes the question subset, and $\mathcal{T}_{t}$ represents semantic templates. The generated prompt $P_{t}$ directs the vision-language model to prioritize question-relevant visual elements while de-emphasizing extraneous details.

We implement a context-aware frame selection strategy that adapts to question scope. If the proportion of high-relevance shots exceeds a threshold $\gamma$, the question is classified as requiring global temporal context. In this case, our VTSearch retrieval function returns representative frames from the breadth-expansion layer, ensuring coverage across the entire video timeline. Conversely, if relevance is concentrated in a few shots, the question is deemed to require local detail. VTSearch then returns representative frames from the depth-expansion layer of high-relevance clips, focusing representational capacity on critical regions.
For the selected frame set $F_{rep}$, the VLM generates targeted captions using the synthesized prompt $P_{t}$, producing descriptions that emphasize question-relevant visual content.
To construct coherent segment-level representations, we aggregate frame-level captions into holistic semantic summaries, which eliminates inter-frame redundancy while preserving factual information, producing segment summaries $S_i^{t}$ that balance comprehensiveness with conciseness.

\subsection{HiCrew: Hierarchical Multi-Agent Framework}

\subsubsection{Knowledge Source Integration and Dynamic Prompting}

HiCrew instantiates a lightweight two-tier reasoning architecture comprising a Planning Layer and an Execution Layer. The Hybrid Tree $T_{hyb}$ serves as a structured, shared knowledge repository accessible to all agents. The Knowledge Source Manager $\mathcal{K}$ implements question-type-aware retrieval, dynamically selecting appropriate semantic descriptions.
Through Retrieval-Augmented Generation, agents can traverse the tree structure to retrieve multi-scale semantic information, from coarse segment summaries to fine-grained frame-level descriptions, ensuring access to both global context and local details as needed.
Furthermore, we implement Question-Type-Aware Dynamic Prompting (QTDP). The system maintains specialized agent configuration files $\mathcal{F}_{t}$ for each question category, encoding targeted reasoning strategies, appropriate tool sequences, and evidence weighting schemes:
\begin{equation}
\mathcal{F}_{t} = \{\text{Strategy}_{t}, \text{Tools}_{t}, \text{Weights}_{t}\}
\end{equation}
where $\mathcal{F}_{t}$ is the agent configuration, comprising reasoning $\text{Strategy}_{t}$, $\text{Tools}_{t}$ sequence, and evidence $\text{Weights}_{t}$. For example, Causal Why (CW) questions employ a Bidirectional Temporal Search Strategy: agents simultaneously search backward for action triggers and forward for action consequences, determining causal directionality through semantic analysis rather than relying solely on temporal proximity.

\subsubsection{Hierarchical Multi-Agent Collaboration}

The Planning Layer serves as the system's cognitive control center, implementing adaptive question orchestration through two specialized agents. The \textit{Problem Analysis Agent} parses the semantic and structural features of input question $Q$, performing two critical functions: 1) question type classification, mapping $Q$ to our predefined taxonomy, and 2) agent role selection, determining the minimal agent subset $\mathcal{A}_{min} \subseteq \mathcal{A}_{all}$ required to answer the question. For instance, static descriptive questions (e.g., "What is the location?") do not require temporal reasoning capabilities, allowing the Visual Analysis Agent to be excluded from the execution workflow, thereby reducing computational overhead without sacrificing answer quality.
The \textit{Task Planning Agent} receives the filtered agent list from the Problem Analysis Agent and synthesizes a structured execution workflow. This workflow specifies agent invocation sequences, task assignments for each stage, and input-output specifications that govern inter-agent communication. By constructing question-specific reasoning paths, this agent enables the system to allocate reasoning resources proportional to question complexity.

The Execution Layer comprises domain-specialized agents operating under the ReAct (Reasoning and Acting) paradigm, executing the workflow orchestrated by the Planning Layer. The \textit{Text Agent} specializes in question-aware caption retrieval and textual reasoning. Equipped with a comprehensive toolchain for multi-granularity retrieval, it first obtains the video's temporal index, then retrieves question-type-specific semantic descriptions for relevant moments, and finally extracts segment-level summaries $S_i^{t}$ to construct a textual evidence base grounded in the Hybrid Tree structure.
The \textit{Visual Analysis Agent} focuses on extracting and verifying dynamic visual information through direct frame analysis. It invokes a VLM to perform deep semantic interpretation of key video segments, adapting its analytical focus based on question type. For causal questions, it emphasizes action-object interactions; for temporal questions, it tracks state changes across frames; for descriptive questions, it provides detailed scene and attribute descriptions.
The \textit{Evidence Integration Agent} performs cross-modal evidence fusion and conflict resolution. It implements a question-type-aware weighting strategy, for example, assigning higher weight ($w_{visual}=0.7$) to visual evidence for descriptive questions while prioritizing textual reasoning for temporal ordering questions. The agent aggregates upstream outputs through weighted combination:
\begin{equation}
\text{Score}_{option} = \sum_{i} w_i \cdot conf_i \cdot \text{Evidence}_i,
\end{equation}
where $conf_i \in [0,1]$ represents the confidence score of evidence source $i$. Additionally, for causal questions, this agent performs bidirectional consistency verification to ensure that identified causes and effects maintain logical coherence.
The \textit{Answer Generation Agent} synthesizes the integrated evidence into a final answer, performing format validation to ensure output falls within the valid option space and is sufficiently supported by evidence. It also provides transparency by generating explanation traces that document the reasoning path from evidence to conclusion.

\section{Experiments}

\subsection{Experimental Setup}
\textbf{Benchmarks.} We evaluate HiCrew on two long-form video understanding benchmarks:
\textit{EgoSchema} \cite{b14} contains over 5,000 multiple-choice questions across 250+ hours of egocentric video. Each question requires selecting from five options based on a 180-second clip, with long temporal horizons challenging coherent reasoning across extended time scales.
\textit{NExT-QA} \cite{b15} evaluates causal and temporal reasoning with 5,440 videos and nearly 52,000 question-answer pairs. Questions are categorized as Causal, Temporal, or Descriptive, enabling detailed performance analysis across reasoning dimensions.

\begin{table}[!t]
\caption{Performance on EgoSchema (subset/full) and NExT-QA (Temporal/Causal/Descriptive). Best and second-best results are \colorbox{best}{red} and \colorbox{second}{green}.}
\centering
\resizebox{\columnwidth}{!}{
\setlength{\tabcolsep}{1pt}
\renewcommand{\arraystretch}{1.2}
\begin{tabular}{lccccccc}
\toprule
\textbf{Model} & \textbf{(M)LLM} & \multicolumn{2}{c}{\textbf{EgoSchema}} & \multicolumn{4}{c}{\textbf{NExT-QA}} \\
\cmidrule(lr){3-4} \cmidrule(lr){5-8}
 &  & \textbf{Sub} & \textbf{Full} & \textbf{Tem} & \textbf{Cau} & \textbf{Des} & \textbf{Avg} \\
\midrule
AKeyS \cite{b23} & GPT-4 & 68.0 & 63.1 & 72.3 & 78.2 & 85.4 & 77.4 \\
VideoTree \cite{b4} & GPT-4 & 66.2 & 61.1 & 70.6 & 76.5 & 83.9 & 75.6 \\
LifelongMemory \cite{b29} & GPT-4 & 64.1 & 58.6 & - & - & - & - \\
VideoAgent \cite{b2} & GPT-4 & 60.2 & 54.1 & 68.3 & 68.3 & 74.3 & 71.3 \\
VideoChat2 \cite{b22} & Mistral-7B & 63.6 & 54.4 & - & - & - & \cellcolor{second}78.6 \\
VideoLLaMA3-7B \cite{b28} & - & 63.3 & - & - & - & - & - \\
LLoVi \cite{b26} & GPT-4 & 57.6 & 50.3 & \cellcolor{second}73.7 & 70.2 & 81.9 & 73.8 \\
LLaMA-VID-7B \cite{b27} & Vicuna-13B & 38.5 & - & - & - & - & - \\
Mobile-VideoGPT \cite{b24} & - & 36.7 & - & - & - & - & 73.7 \\
InternVideo \cite{b25} & VideoChat2-HD-F16 & - & 60.0 & - & - & - & - \\
LVNet \cite{b30} & GPT-4o & 68.2 & 61.1 & 65.5 & 75.0 & 81.5 & 72.9 \\
AKeyS \cite{b23} & GPT-4o & \cellcolor{second}68.6 & \cellcolor{second}63.6 & 72.9 & \cellcolor{second}79.0 & \cellcolor{second}86.1 & 78.1 \\
\midrule
\textbf{HiCrew} & \textbf{GPT-4o} & \cellcolor{best}\textbf{71.6} & \cellcolor{best}\textbf{64.6} & \cellcolor{best}\textbf{74.3} & \cellcolor{best}\textbf{80.4} & \cellcolor{best}\textbf{87.1} & \cellcolor{best}\textbf{79.5} \\
\bottomrule
\end{tabular}
}
\label{tab:main}
\end{table}
\textbf{Implementation Details.} We employ GPT-4o \cite{b16} as the backbone LLM for all main experiments, ensuring strong reasoning capabilities. Following VideoTree \cite{b4}, we utilize EVACLIP-8B \cite{b18} as our visual encoder for feature extraction. For caption generation, we adopt BLIP-2 \cite{b17} for EgoSchema and LaViLa \cite{b19} for NExT-QA, leveraging their respective strengths for different video domains.
Videos are preprocessed with uniform temporal sampling at 1 FPS. For Hybrid Tree construction, we first apply shot boundary detection to segment videos into temporally coherent clips. These clips are then scored by GPT-4o for question relevance, with high-scoring clips (threshold $\tau=2.5$) undergoing selective deep expansion via hierarchical clustering (using $K=2$ for K-Means and a maximum depth of 3). Empirical analysis on validation data indicates that performance remains stable across $\tau \in [2.0, 3.0]$, demonstrating robustness to this hyperparameter. For Question-Aware Captioning, the global context threshold $\gamma$ is set to 40\%, determining when to retrieve from breadth versus depth layers. Regarding computational overhead, queries converge in an average of 1.3 rounds (with a hard maximum of 15 iteration rounds) to ensure convergence while maintaining efficiency.

\textbf{Evaluation Metrics.} We adopt accuracy as the primary metric for EgoSchema, consistent with standard practice. For NExT-QA, beyond overall accuracy, we report performance on each question type (Causal, Temporal, Descriptive) to provide granular insight into the model's reasoning capabilities across different dimensions.

\subsection{Main Results}

\begin{figure*}[!t]
\centering
\includegraphics[width=0.97\textwidth,height=0.62\textwidth]{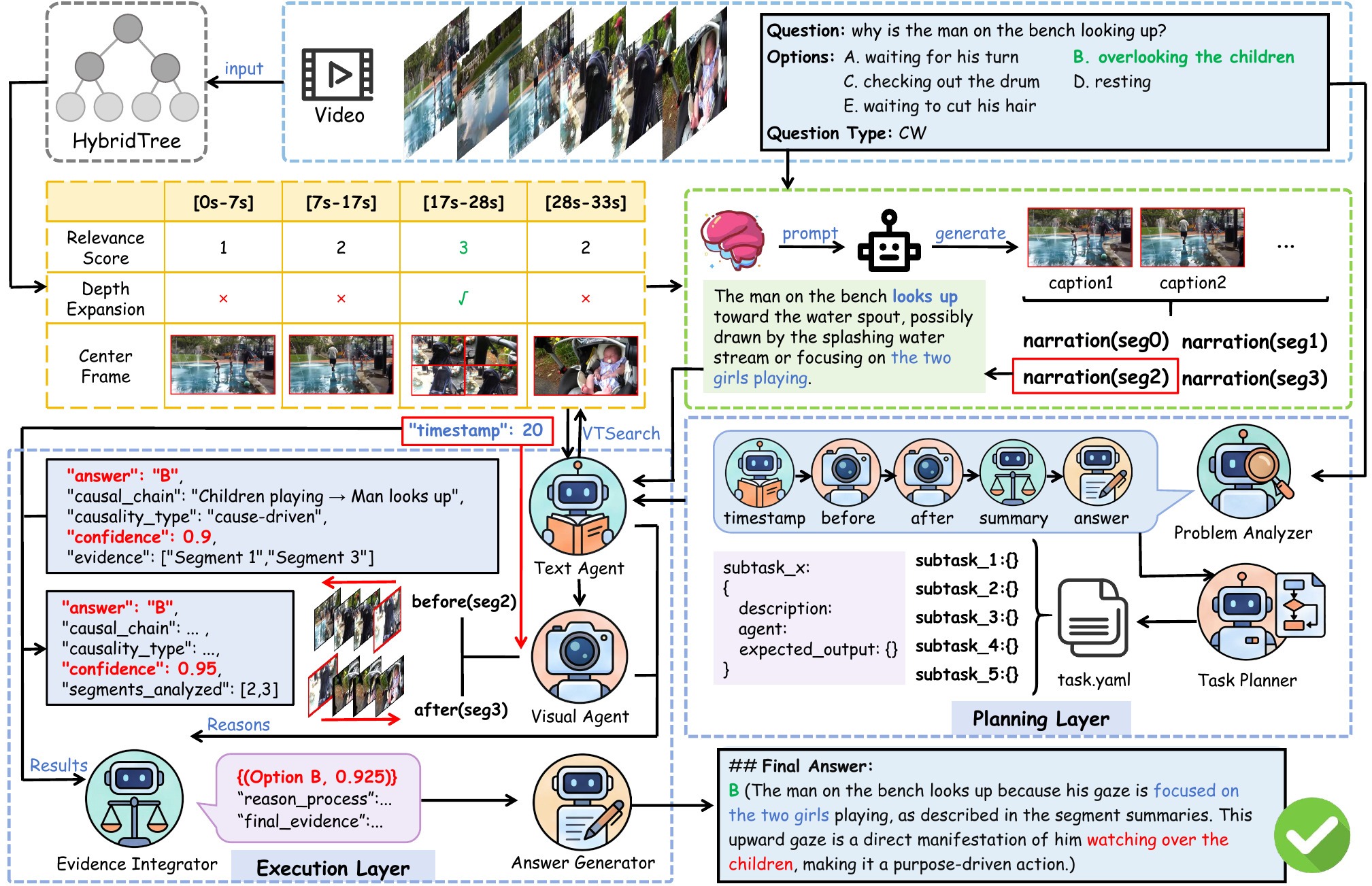}
\caption{Qualitative analysis on a Causal reasoning question from NExT-QA.}
\label{fig:3}
\end{figure*}

Table~\ref{tab:main} presents comparative results across both benchmarks. On EgoSchema, HiCrew achieves 71.6\% accuracy on the subset and 64.6\% on the full set, establishing new state-of-the-art performance. Notably, our method outperforms the structured baseline VideoTree by a substantial margin of 5.4\% on the subset, validating our hypothesis that preserving temporal topology through shot-based initialization provides critical advantages over global clustering approaches that may disrupt narrative coherence.
On NExT-QA, HiCrew attains an average accuracy of 79.5\%, with strong performance on Causal questions (80.4\%). This represents a 1.4\% improvement over the previous best GPT-4o baseline (AKeyS with GPT-4o at 78.1\%), demonstrating the effectiveness of our bidirectional temporal search strategy and dynamic planning mechanisms for handling complex reasoning scenarios. The consistent improvements across all three question types (Temporal: 74.3\%, Causal: 80.4\%, Descriptive: 87.1\%) indicate that HiCrew's adaptive orchestration successfully tailors reasoning strategies to diverse question demands.
Importantly, HiCrew consistently surpasses the recent agent-based method AKeyS across both benchmarks under an identical GPT-4o backbone. This controlled comparison isolates the contributions of our hierarchical collaboration architecture and question-aware mechanisms, demonstrating that our innovations in structured representation and adaptive planning provide genuine advantages beyond simply leveraging more capable foundation models.

\begin{table}[!t]
\caption{Ablation study. $\Delta$ denotes performance drop.}
\centering
\small
\begin{tabular}{lcc}
\toprule
\textbf{Method} & \textbf{Acc. (\%)} & \textbf{$\Delta$} \\
\midrule
\textbf{HiCrew (Full)} & \textbf{71.6} & - \\
w/o Hybrid Tree & 58.7 & -12.9 \\
w/o Q-Aware Captioning & 59.2 & -12.4 \\
w/o Planning Layer & 62.0 & -9.6 \\
\bottomrule
\end{tabular}
\label{tab:ablation}
\end{table}
\subsection{Ablation Study}
To systematically understand the contribution of each component, we conduct controlled ablation experiments on the EgoSchema subset, progressively removing key innovations while maintaining all other implementation details.
Replacing $T_{hyb}$ with uniform frame sampling results in a dramatic 12.9\% performance drop, underscoring the importance of structured representation that preserves temporal topology. Uniform sampling fails to capture hierarchical organization and discards fine-grained details in question-relevant segments. Reverting to question-agnostic captions decreases accuracy by 12.4\%, demonstrating that caption quality is as crucial as structural representation. Generic captions frequently omit visual details essential for specific queries, while our intent-driven mechanism actively directs the VLM's attention to question-relevant elements. Removing adaptive orchestration in favor of a fixed agent workflow yields a 9.6\% reduction, validating the value of question-specific reasoning strategies. The fixed workflow either applies insufficient reasoning to complex questions or wastes computation on simple ones.
These ablation results validate that each component contributes meaningfully to overall performance, with their combination creating synergistic effects.

\subsection{Qualitative Analysis}

Fig.~\ref{fig:3} illustrates HiCrew's reasoning process on a challenging NExT-QA causal question: \textit{Why is the man on the bench looking up?} The video is segmented into four shots ($s_1$-$s_4$) via shot boundary detection. Relevance assessment identifies $s_3$ as highly pertinent, triggering deep hierarchical expansion, while low-relevance shots ($s_1$,$s_2$,$s_4$) remain as leaf nodes. This selective strategy achieves efficient compression while preserving temporal topology for causal reasoning.
The intent analysis module classifies this as a Causal question and synthesizes a targeted prompt $P_{causal}$ instructing the VLM to describe triggers and contextual factors. The resulting caption captures \textit{children playing near a fountain}, providing essential environmental context that generic captions miss. The Planning Layer initiates a bidirectional temporal search workflow where the Text Agent retrieves the question-aware caption, the Visual Analysis Agent verifies spatial consistency between the children's location and the man's gaze direction, and the Evidence Integration Agent synthesizes these cues to select the correct answer: \textit{overlooking the children} (Option B).
In contrast, VideoTree disrupts temporal structure through global clustering and misses critical environmental context, incorrectly inferring the man is \textit{looking at a bird.} This demonstrates how HiCrew's integrated approach enables robust causal reasoning beyond simple pattern matching.

\section{Conclusion}
We have presented HiCrew, a hierarchical multi-agent framework for long-form video understanding. Our Hybrid Tree structure preserves temporal topology through shot boundary detection while achieving effective compression via selective hierarchical clustering. The Question-Aware Captioning mechanism generates intent-driven semantic descriptions aligned with question-specific demands. The Planning Layer enables adaptive orchestration by dynamically selecting agent roles and reasoning paths based on question complexity. Extensive experiments on established datasets validate the effectiveness of our approach, with particularly strong performance in temporal and causal reasoning tasks.

\bibliographystyle{IEEEbib}
\bibliography{icme2026references}

\end{document}